\documentclass[9pt]{article}

\usepackage{amssymb}

\usepackage[margin=1in]{geometry}

\usepackage{longtable}
\usepackage{lscape}
\usepackage{array}

\usepackage{caption} 

\usepackage{supertabular,booktabs}

\usepackage{cite}

\usepackage[figuresright]{rotating}

\usepackage{xcolor}
\usepackage{adjustbox}

\usepackage{graphicx}

\usepackage{amsmath}

\usepackage{wrapfig}
\usepackage{subfigure}

\title{Detection of Opioid Users from Reddit Posts via an Attention-based Bidirectional Recurrent Neural Network}

\author{Yuchen Wang$^{\text{1,*}}$, Zhengyu Fang$^{\text{1,*}}$, Wei Du$^{\text{1}}$, Shuai Xu$^{\text{1}}$, Rong Xu$^{\text{2}}$,  and Jing Li$^{\text{1,\S}}$\\
$^{\text{1}}$Department of Computer and Data Sciences, $^{\text{2}}$Center for Artificial Intelligence\\ 
in Drug Discovery, Case Western Reserve University, Cleveland, 44106, USA}

\begin{document}

\date{}

\maketitle

\begin{abstract}
The opioid epidemic, referring to the growing hospitalizations and deaths because of overdose of opioid usage and addiction,  has become a severe health problem in the United States. Many strategies have been developed by the federal and local governments and health communities to combat this crisis. Among them, improving our understanding of the epidemic through better health surveillance is one of the top priorities. In addition to direct testing, machine learning approaches may also allow us to detect  opioid users by analyzing data from social media because many opioid users may choose not to do the tests but may share their experiences on social media anonymously. In this paper, we take advantage of recent advances in machine learning,  collect and analyze user posts from a popular social network Reddit with the goal to identify opioid users. Posts from more than 1,000 users who have posted on three sub-reddits over a period of one month have been collected. In addition to the ones that contain keywords such as opioid, opiate, or heroin, we have also collected posts that contain slang words of opioid such as black or chocolate. We apply an attention-based bidirectional long short memory model to identify opioid users. Experimental results show that the approaches significantly outperform competitive algorithms in terms of F1-score. Furthermore, the model allows us to extract most informative words, such as opiate, opioid, and black, from posts via the attention layer, which provides more insights on how the machine learning algorithm works in distinguishing drug users from non-drug users. \\
\end{abstract}

\textbf{CCS CONCEPTS} • Bioinformatics • Natural language processing • Web searching and information discovery

\textbf{Additional Keywords and Phrases}: Social network mining, Opioid user detection, Attention-based bidirectional long short memory model

\section{Introduction}
Opioids are a class of prescription medication for treating chronic and acute pain. However, opioids have many side-effects (e.g., hypoventilation), negatively impact the physical and mental health of their users, and  may cause severe addiction even under doctors' guidance. From 2013 to 2017, the number of opioid-involved overdose deaths increased from 25,052 to 47,600 \cite{GladdenRMODonnellJMattsonCL2019}. Moreover, approximately 3.3\% of patients, who are exposed to chronic opioid therapy, will become addicted \cite{Kalkman2019}. In addition, there are also millions of Americans addicted to opioids due to illegal usage of drugs without prescriptions. Opioid addiction has become one of the most severe epidemics in the US. 

To understand better the problem of opioid addition, further research is needed. In particular, the role of social media in studying opioid epidemic has become increasingly important, mainly because of the large number of social media users, including opioid users, and the large volume of data posted on social media at any time point. For many opioid users, they may not be willing to tell the true story or any details about their addiction to their friends or family members. However, they may sometimes share their feelings or ask for help on social media anonymously. it is not uncommon that users choose to share their experience, or seek for help on the social media platforms such as Reddit or Twitter (now X). Others such as drug dealers may also use social media to sell their products. Therefore, data on social media can provide a great deal of information that can be supplementary to the knowledge of domain professionals and help us to gain new insights into opioid addiction.

In recent years, there had been an increasing trend in utilizing social media data to better understand health related problems in general. Earlier work focused on detecting and extracting adverse drug reactions (ADRs) \cite{Yates2013,Liu2016} based on users' comments on social media. More recently, several groups have developed different approaches to address problems associated with opioid epidemic. Via a series of papers \cite{Zhang2016,Fan2017,Fan2018}, the authors have established a heterogeneous information network (HIN) based model called AutoDOA for opioid addiction detection based on Twitter data. Their research focused on the mining of relationships among users and relationships among tweets from different users. Mackey et al. \cite{Mackey2017} utilized topic modeling method to detect  illegal online sales of controlled substances, which was also based on Twitter data. However, data from tweets has its own limitations due to their limited length. Yang et al. \cite{Yang2018} utilized posts from Reddit, which could be much longer, in predicting opioid relapse. They have developed a generative adversarial network (GAN) based approach to predict the addiction relapses based on sentiment images and social influences. Other researchers \cite{Yao2020} have utilized Reddit data to detect suicidality among opioid users. 

Although the analysis of opioid addiction based on social media data draws some attention in recent years, further research is needed, especially for opioid user detection. It is of great significance to detect opioid users so some help may be provided to them down the road. Trained health professionals and social workers can potentially provide help to the likely opioid users by answering their questions anonymously, or provide further interventional assistant with the permission from the users. Results can also be used for other research topics such as predicting opioid relapse. The objective of our research is to detect opioid users accurately based on the social media data from Reddit. Unlike the previous work that was mainly focusing on relationships from Twitter users, we utilize a deep learning approach by directly analyzing user posts from Reddit. As mentioned earlier, information obtained from Twitter is limited because of the constraint on the number of characters, and they can be highly noisy due to lack of feedback or monitoring. Reddit is a community based social media for discussion topics with content curated by the community through voting, which can generate more reliable information for us to learn insights on the opioid epidemic. 

In this paper, we apply an Attention-based Bidirectional Long Short Term Memory (Att-BLSTM) model~\cite{Zhou2016,Ma2017,LI2020} to identify opioid users accurately from Reddit posts. Long short-term memory (LSTM)~\cite{Hochreiter1997,Cheng2016} is an artificial recurrent neural network (RNN) architecture that consists of feedback connections. It can process static data such as images, but more often it is applied to sequence data such as speech, text, or video. BLSTM, rooted from Bidirectional RNN, allows the model to get information from the past and the future simultaneously. Att-BLSTM further utilizes neural attention mechanism to capture the most important semantic information in a sentence. The framework has been widely used  in other applications. \cite{Zhou2016} utilized Att-BLSTM to handle the relation classification problem and demonstrated that Att-BLSTM  could achieve good performance using raw text instead of lexical resources.  \cite{Wang2016} applied Att-BLSTM to sentiment classification.  \cite{Seo2016} proposed a hierarchical attention network for document classification with  two levels of attention mechanisms applied at the word-level and the sentence-level. 

We first obtain our data by crawling Reddit using the APIs provided by its platform, focusing on three sub-topics including ``opiates", ``drugs" and ``opiatesRecovery". Posts from more than 1,000 users over a period of one month have been collected. In addition to the ones that contain keywords such as opioid, opiate, or heroin, we have also collected posts that contain slang words of opioid such as black or chocolate. All the users and their posts are manually reviewed via crowdsourcing by student researchers with the guidance of the senior researchers on the study, and each Reddit user is labeled by at least two students. The Att-BLSTM model with three variants are compared with the baseline BLSTM model without the attention layer, as well as five commonly used classification methods such as SVM and Random Forests. Experimental results show that Att-BLSTM methods significantly outperform competing algorithms. Furthermore, the model allows us to extract most informative words, such as opiate, opioid, and black, from posts via the attention layer, which provides more insights on how the machine learning algorithm works in distinguishing drug users from non-drug users. Our main contributions are summarized as follows:
\begin{itemize}
	\item This work provides a novel application of Att-BLSTM to solve the opioid user detection problem from Reddit. The model is suitable for handling long posts in our dataset and outperforms many existing methods.
	\item A systematic and comprehensive experiment is implemented using a real-world dataset. With the help of student researchers, we create a manually curated dataset that not only contains keywords, such as ``opium", ``opiates", but also contains slang words for opioid, such as  ``black", ``chocolate". This expands the application scenarios of the model. The anonymous dataset, which is available at https://github.com/jinglicase/OpioidUserDetection, can also serve as a resource to test future methods.  
	\item The most significant words in each post can be visualized via the attention layer, which can improve the  explainability of the prediction results from the machine learning method. 
\end{itemize}

\section{Methods}
\begin{wrapfigure}{R}{0.5\textwidth}
\centering
\includegraphics[scale=0.2]{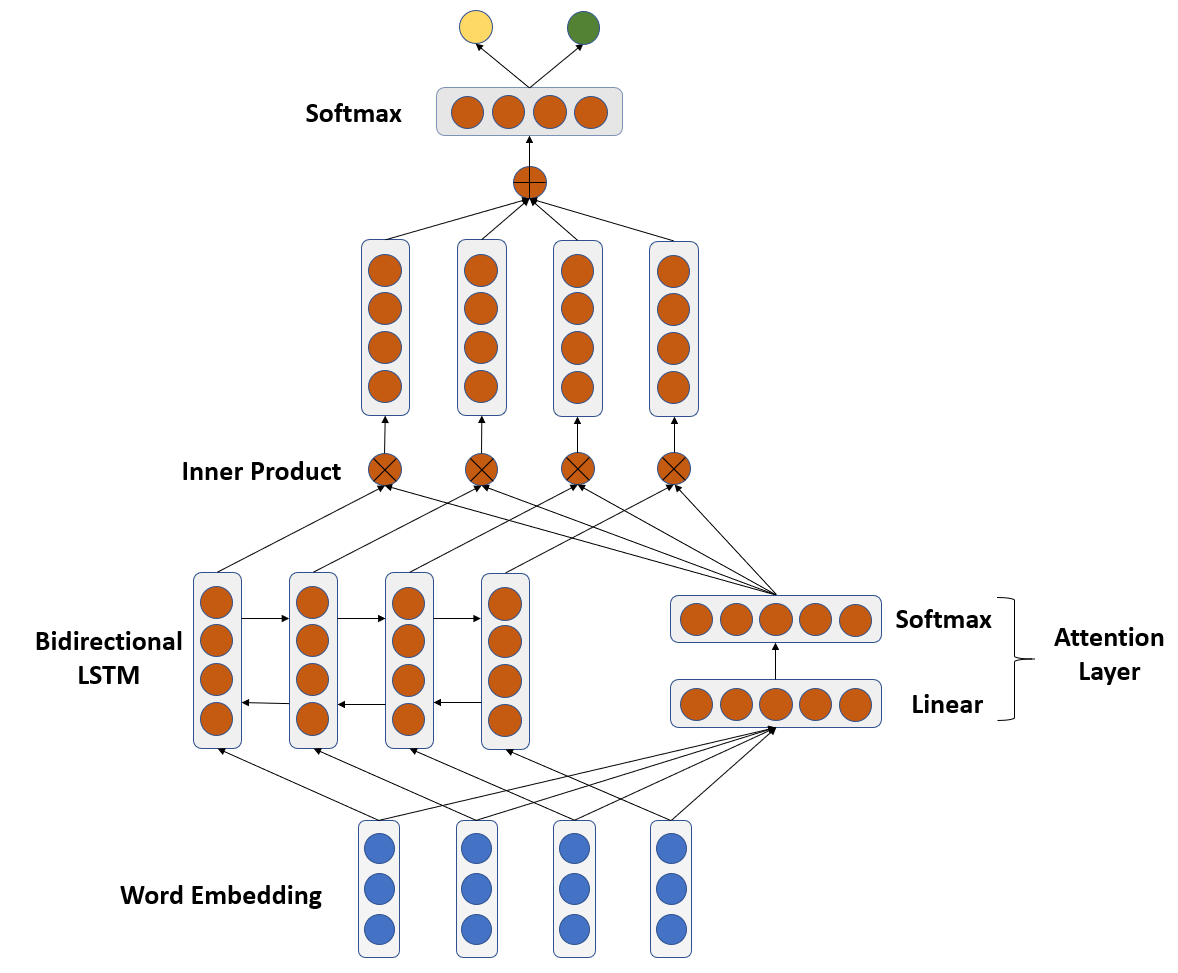}
\caption[The structure of the Att-BLSTM model]{The structure of the Att-BLSTM model, consisting of four main substructures: a word embedding layer, a bidirectional LSTM layer, an attention layer, and a fully connected layer. \label{network_structure}}

\end{wrapfigure}
%
The structure of the Att-BLSTM model is shown in Figure~\ref{network_structure}. It consists of four main substructures:  a word embedding layer, a bidirectional long short term memory layer, an attention layer, and a fully connected layer. The word embedding layer converts words into vector representations to better capture their similarities. The BLSTM layer allows incorporation of an increased amount of input information from both directions, which is desirable in our application  given that the posts from Reddit could be very long.  Furthermore, it overcomes gradient vanishing problem via an adaptive gating mechanism. The attention layer generates the weights for the words and adjusts the significance of words utilizing these weights, which can be used to highlight the critical portion of posts in predicting drug users and to make the model more explainable. The fully connected layer with a sigmoid activation function is used to identify/classify opioid drug users. We briefly introduce each layer in the sequel. 
\subsection{Word embedding}
The word embedding layer maps each word in a post into a real-valued vector representation, which allows words with similar meanings to have similar representations. The parameter to be learned is the word embedding matrix $M_{\textit{d} \times \textit{v}}$, where $v$ is the size of a fixed vocabulary and $d$ is a hyper-parameter indicating the size of the embedding. The parameters can be initialized \textit{randomly} and can be updated iteratively during the training. In some cases, the word embedding matrix $M_{\textit{d} \times \textit{v}}$ can be learned \textit{beforehand} based on some large corpus and can be used directly in other applications. The values in $M$ will therefore be kept constant during the training. In the third scenario, pre-trained parameters will be used as initialization values but will be \textit{updated} during training. We will consider all three variants in our experiments. To obtain a pre-trained word embedding matrix, we will use the dataset obtained by applying the Global Vectors for Word Representation (GloVe) algorithm on a large number of Wikipedia  pages in 2014 \cite{Glove}. GloVe is an unsupervised learning algorithm  for obtaining vector representations of words based on aggregated global word-word co-occurrence statistics from a corpus. The dataset contains four different pre-trained word embedding matrices with different values for the dimension hyper-parameter  $d$: 50, 100, 200, and 300. In our experiment, we will test the performance of our approach using all these four values.  

\subsection{BLSTM architecture}
LSTM~\cite{Hochreiter1997,Cheng2016} is an artificial recurrent neural network (RNN) architecture that consists of feedback connections, which can be applied to sequence data such as speech, text, or video. BLSTM~\cite{Schuster1997}, rooted from Bidirectional RNN, consists of two parallel LSTM units in the opposite directions, which allows the model to get information from the ``past'' and the ``future'' simultaneously. LSTM~\cite{Hochreiter1997} was proposed initially to address the vanishing gradient problem, by utilizing an adaptive gating mechanism. LSTM architecture consists of a set of recurrently connected subnets called memory blocks. Each block contains three gates: an input gate $i_t$, an output gate $o_t$, and a forget gate $f_t$. In addition, each block is also associated with a hidden state $h_t$, a cell state $c_t$, and a cell input activation $\tilde{c}_t $. The updating of the gates, states of the current cell is based on the current input, previous hidden state, and cell state via the following formula. 
\begin{align}
f_t &= \sigma(W_f x_t + U_f h_{t-1} + b_f) \\
i_t &= \sigma(W_i x_t + U_i h_{t-1} + b_i) \\
\tilde{c}_t &= tanh(W_c x_t + U_c h_{t-1} + b_c) \\
c_t &= i_t*\tilde{c}_t + f_t*c_{t-1} \\
o_t &= \sigma(W_o x_t + U_o h_{t-1} + b_o) \\
h_t &= o_t*tanh(c_t)
\end{align}
\noindent where $x_t \in R^d$ represents the input vector of the unit at time $t$ (\textit{i.e.}, the embedding of $t^{th}$ word in an input post for our application), $W_k  \in R^{l \times d}$ and $U_k  \in R^{l \times l}$ denote the weight matrices,  $b_k \in R^l$ are the bias vectors,  $k \in \{f, i, c, o\}$,  $l$ is the number of hidden units which is a hyperparameter, and  $\sigma$ is the sigmoid function. The punctuation $*$ represents pairwise product of two vectors. 
BLSTM consists of two LSTM layers of different directions. The final output of BLSTM is just the concatenation of the two hidden state vectors $h_t = [\overrightarrow{h_t}, \overleftarrow{h_t}]$.

%
\subsection{Attention layer}
Neural networks with attention layers are widely used in image captioning, language translations, speech recognition among other applications. In our method, the attention layer is utilized to obtain a weight for each input word, indicating the relative importance/contribution of the word to the classification task. Let $X \in R^{d  \times T}$ denote the input matrix where $d$ is the length of  word vectors and $T$ is the length of posts.  
The output of the attention layer $\alpha$ can be calculated as: 
\begin{align}
\alpha &= softmax(w_{\alpha}^T X + b_{\alpha})
\end{align}
\noindent where  $w_{\alpha} \in R^d$ is the weight vector and $b_{\alpha} \in R^T$ is the bias vector.


\subsection{Classification layer}
The classification unit consists of one fully connected layer, which takes the dot product of the BLSTM layer and the attention layer as its input, and uses a SoftMax layer to make final predictions.  


\subsection{Model variants}
Because there are different ways of initializing the model parameters in the word embedding layer, we analyze three variants of the Attention-Based BLSTM (Att-BLSTM) model. For Att-BLSTM-M1, the word-embedding vectors are initialized randomly and will be learned from our own data; for Att-BLSTM-M2, the word-embedding vectors are initialized using the pre-trained parameters based on the GloVe dataset and are fixed during the training process; and for Att-BLSTM-M3, they are initialized using the pre-trained parameters based on GloVe dataset but are updated during the training process. In addition to the Att-BLSTM model and its variants, we will also consider the basic  BLSTM model without the attention layer and its corresponding three variants based on different initialization strategies:  BLSTM-M1, BLSTM-M2, and BLSTM-M3. 

%
%
	
\section{Experiments}

\subsection{Data Collection and Labeling}
Our data was collected from Reddit using Python Reddit API Wrapper (PRAW), a web crawling tool. The data was obtained from three subreddits called `Opiates', `OpiatesRecovery', and `Drugs'. For each subreddit, we collected the comments of redditors who were in the ``Top-Month" tab from January 6th 2020 to February 5th 2020. The collection was based on opioid-related keywords (e.g., opium, opioid) as well as their slang words (e.g., black, chocolate) obtained from ``Drug Slang Code Words" in the DEA Intelligence Report \cite{Dea2017}. Comments with less than 15 words were excluded.  In total, comments from 1058 redditors were collected. All comments from the same redditor were concatenated into one ``post'', representing the user. Therefore, in this paper, we use ``post'' and ``user'' interchangeably. All user identification information was removed before further analysis to protect user privacy.  

The initial dataset  had no labels. The label of each user/post was obtained manually by crowdsourcing with the help of student researchers. Each post was read by two students independently and was assigned to one of five possible labels. If the labels from the two students were the same, that label was used for the post/redditor. Otherwise, a third student read it again and provided her/his own result. The label with two votes was deemed the correct one. The five labels are: non-opioid users, opioid users, previous opioid users (who actually quit now),  drug dealers, and users who use prescription drugs. Their corresponding numbers of posts/redditors are  410, 471, 107,  27, 43, respectively. The total number of cases that the first two students disagreed is 90 (8.5\% of the total number of redditors), which illustrates the inherent difficulty of the problem itself.  Because the total number of redditors in the last three categories was relatively small, in the current study, we only focused on distinguishing opioid users from non-opioid users. 

\subsection{Data Preprocessing}
For data preprocessing, we converted all letters to lowercase, removed all punctuation marks and stop words, and lemmatized all posts. 
Among these 881 users, 254 of them mentioned both opium related keywords as well as their slangs. The numbers of users mentioned only keywords or only slang words are 241 and 386, respectively. The total number of occurrences of each keyword or slang word from all the posts are shown in Table~\ref{tab:numkeySlangWords}. 

\begin{table}[h]
	\centering
	\caption{The number of times of different keywords (in red) and slang words (in black) mentioned in the collected dataset.}
	\label{tab:numkeySlangWords}
	\begin{tabular}{|c|c|c|c|c|c|c|c|}
		\hline
		\color{red}opiates  & black & \color{red}opiate & \color{red}opioid & dreams & \color{red}opium & chocolate & china\\ \hline
		766 & 522 & 447 & 265 & 150 & 97 & 60 & 46 \\ \hline
		gum & toys & incense & pox & hops & cruz & auntie & hocus \\ \hline
		27 & 26 & 4 & 4 & 3 & 3 & 2 & 1 \\ \hline
	\end{tabular}
\end{table}

\begin{figure}
    \centering
    \subfigure[]{\includegraphics[width=0.4\textwidth]{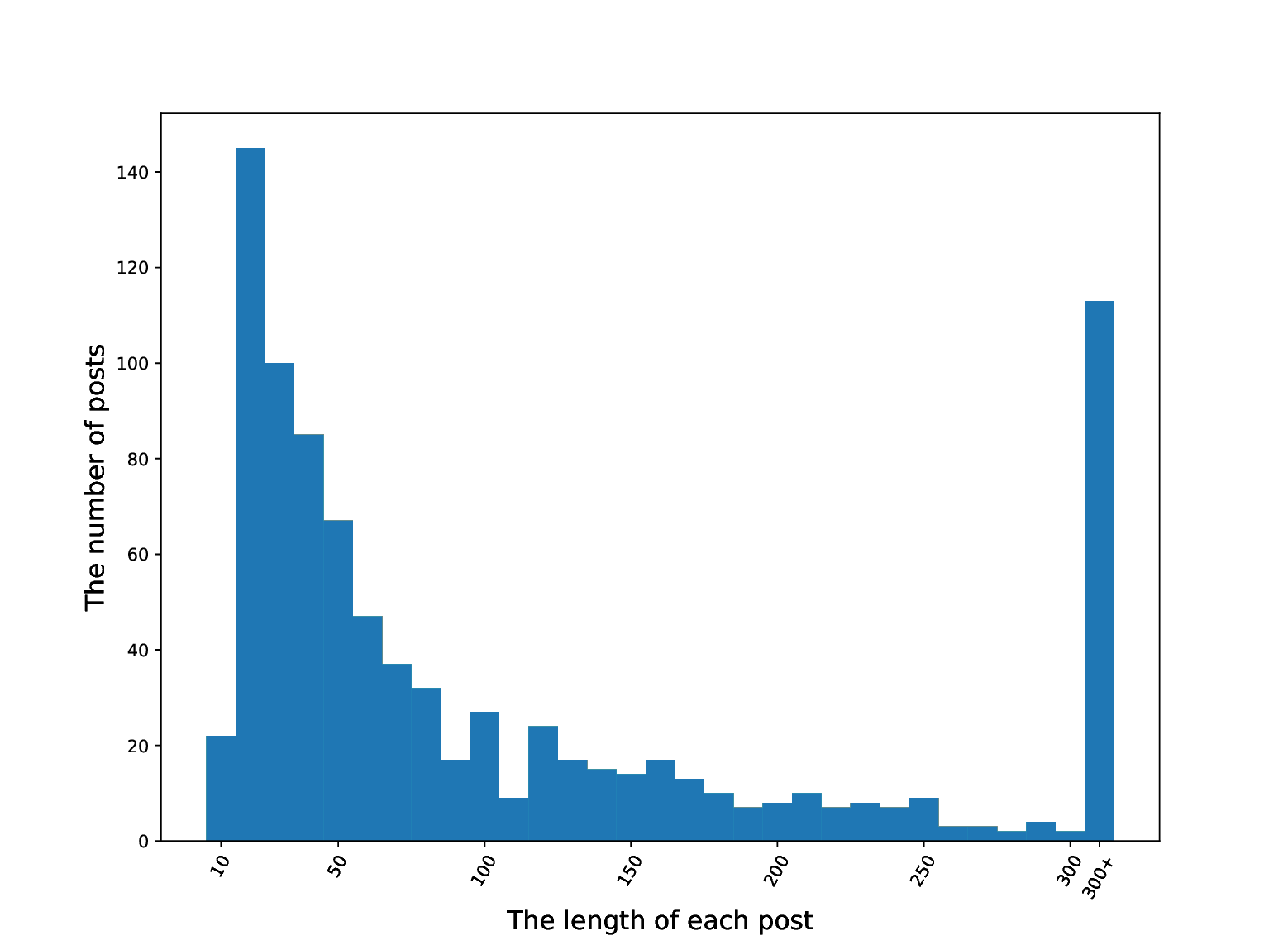}} 
    \subfigure[]{\includegraphics[width=0.58\textwidth]{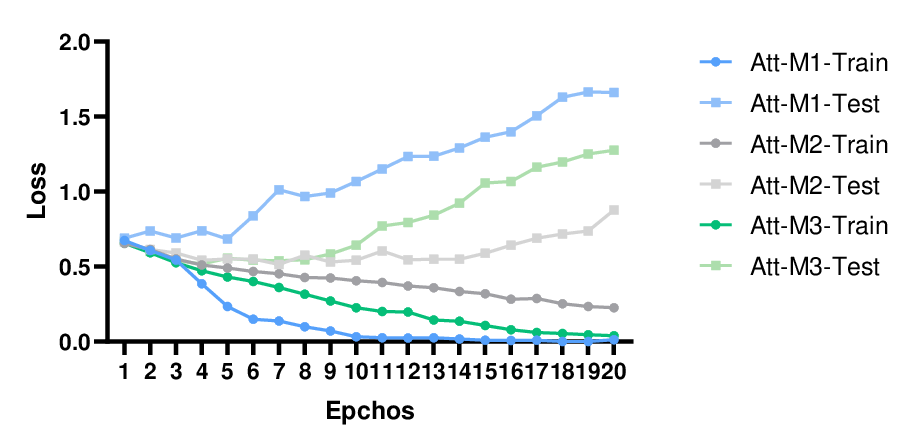}} 
    \subfigure[]{\includegraphics[width=0.58\textwidth]{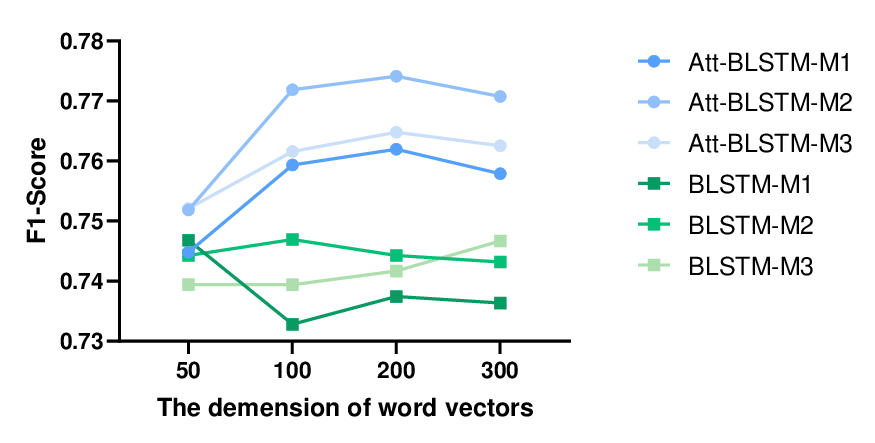}}
    \subfigure[]{\includegraphics[width=0.4\textwidth]{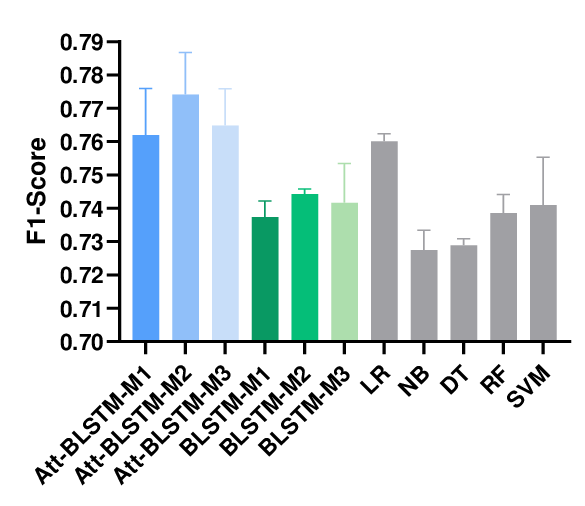}}
    \caption{(a) The length distribution of all posts. (b) The loss on training and testing datasets during training process. (c) The impact of the dimension of word vectors. (d) Comparison of different approaches in terms of F1-Score.}
    \label{fig:combined}
\end{figure}

The length distribution of all posts after preprocessing is shown in Figure~\ref{fig:combined}.a. Although in theory, models based on RNN can handle sequences of arbitrary lengths, many implementation frameworks, such as Keras that we used in this work, expect a fixed length input. 
Based on the length distribution (Figure~\ref{fig:combined}.a),  nearly 70\% of all posts in our collection have less than 100 words. Therefore, we selected the length of 100 words for each post. Longer posts were truncated to get their first 100 words and shorter posts were padded with ``$<$END$>$" at the end. 


\subsection{Experimental Design}
Our models have two important hyperparameters:  the number of epochs in training the models and the dimension of word vectors as the inputs. We will first study the impact of the number of epochs on training and testing errors to determine an appropriate number to avoid overfitting. For the dimension of word vectors, because there are four different dimensions (50, 100, 200, 300) in the GloVe dataset, we will test our models based on all four dimensions.  


Although opioid addiction attracts increasing attention these years, detecting opioid users based on social media data is still a novel topic and existing approaches are limited. Therefore, we will compare our models against several classical baseline approaches, including Logistic Regression, Naive Bayes, Decision Tree, Random Forest, and Support Vector Machine (SVM). We will utilize the implementations in Sci-kit  Learn library and all the parameters for these classifiers are set to be default. The posts will be represented based on the bag-of-words. We will apply the five-fold cross validation method to obtain their prediction results. 
To compare the performance of different approaches, we use the standard metrics including accuracy and F1-score,  with F1-score as the main metric because it represents a balance between precision and recall. Furthermore, we will use Wilcoxon signed-rank test to assess the significance of performance differences among different methods.

Finally, by taking advantage of the attention layer in our models, we can visualize the importance of different words in the whole dataset as well as  in individual posts. Doing so can potentially allow us to depict the relationship between important words and prediction results, leading to an explainable machine learning model. In addition, we can also assess the importance of slang words in distinguishing drug users from non-users. 
\section{Results}
\subsection{Hyper Parameters}


The objective function in training our model is the binary cross-entropy loss and we use a gradient descent optimization algorithm (i.e., AdaMax) for training. We first explored the impact of the two hyperparameters: the number of epochs and the length of word vectors. In our experiments, we first separated our dataset into training (80\%) and testing (20\%) data. Among the training data, we further separated the data into actual training (70\%) and a validation set (30\%). Overall, we used 56\% for training, 24\% for validation, and the rest 20\% for final test. To assess the impact of the number of epochs, we fixed all other parameters (e.g., length of word vector equals 100) and examined the training loss and the validation loss for different number of epochs. Results using BLSTM and Att-BLSTM 
show that the loss of the training process decreased continually while the validation loss  decreased at the beginning and then increased quickly after a few iterations for all model variants,  indicating an overfitting of the models after several  iterations (Figure~\ref{fig:combined}.b). Therefore, in our experiments we set the number of iterations to be five.

Another hyperparameter, which may have influences on the prediction, is the dimension of word vectors. Since there are four kinds of dimensions (50, 100, 200, 300) in GloVe dataset, we also did our experiment based on these four dimensions. The effect of different dimensions on the results is shown in Figure~\ref{fig:combined}.c using the F1-score metric. It seems not sufficient for the Att-BLSTM models to use a dimension of 50. However, for each of the model variants, the performance is fairly stable when the dimension vary from 100 to 300.  
We choose the dimension of 200 to do our experiment when comparing with other baseline methods.

%

\subsection{ Prediction Results}
The performance of  Att-BLSTM and its variants, as well as the baseline approaches, based on three runs of 5-fold cross-validation experiment is shown in Figure~\ref{fig:combined}.d. Clearly,  Att-BLSTM and its variants outperform all other approaches (BLSTM, Logistic Regression, Naive Bayes, Decision Tree, Random Forest, and SVM) in terms of F1-score. In particular, the model using the fixed pre-trained word-embedding parameters (Att-BLSTM-M2) achieves the best F1 score. The other two variants (Att-BLSTM-M1 and Att-BLSTM-M3) have similar result. Notice that by adding the attention layer, variants of Att-BLSTM have a better performance than variants of BLSTM. To assess the statistical significance of the performance difference, we perform Wilcoxon signed-rank test to compare the average performance of the corresponding variants of  Att-BLSTM and BLSTM. Results show that all performance differences are statistically significant with $p$ values of 0.0353, 0.00632, 0.0479 for Att-BLSTM-M1 vs BLSTM-M1,  Att-BLSTM-M2 vs BLSTM-M2, and Att-BLSTM-M2 vs BLSTM-M2, respectively. Variants of BLSTM outperform all other approaches with the exception of Logistic Regression, which is only inferior to Att-BLSTM models.


\subsection{Visualization of Attention Layer}
%
\begin{wrapfigure}{R}{0.4\textwidth}
\centering
\includegraphics[scale=0.5]{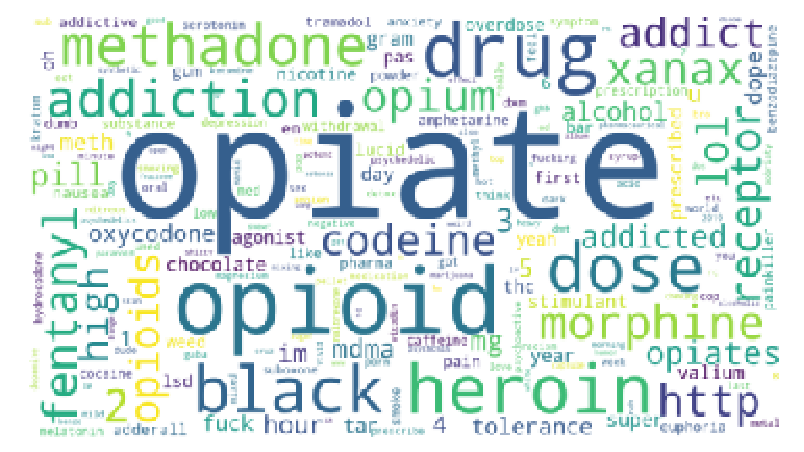}
\caption[The word cloud of the five most important words from each post]{The word cloud of the five most important words from each post, which clearly illustrates that Att-BLSTM can capture important words in distinguishing drug users from non-users. \label{wordCloud}}

\end{wrapfigure}
Results earlier show that with the attention layer, our models achieve better performance. Parameter weights on the attention layer may provide additional information to allow us to understand how and why the model works. To show this, for each post, we select  five words that receive the highest weights from the attention layer and combine all such words together to establish a dictionary. We record the number of times that each word is selected. The more times a word is selected, the more important  it is in distinguishing opioid users from non-users. We create a word cloud based on the dictionary (Figure~\ref{wordCloud}).  Clearly, many drug names such as ``opiate'', ``opioid'', ``heroin''  are all on the top of the list. Interestingly, some slang words such as  ``black'', ``chocolate'' are also on the list, indicating that including slang words provides important information for the model to distinguish drug users from non-users. In addition, it makes intuitive sense  that some other words such as  ``dose'',  ``tolerance'', ``addiction''  are also overly represented. Although not every word is obviously related to opioid addiction, most of the selected words have a close relationship with the task of predicting opioid users. The figure clearly demonstrates that the attention layer actually gives more weights to opioid-related words, which explains why Att-BLSTM  outperforms BLSTM.

To further illustrate what are the important words in a post our model utilizes in separating opioid users from non users, we select and visualize four individual posts in Table~\ref{tab:attention}: one  from an opioid user containing the keyword ``opiates'', one  from a non opioid user containing the keyword ``opiates'',  one  from an opioid user containing the slang word ``china'', and one  from a non opioid user containing the slang word ``china''. It clearly shows that the model was able to capture many important words, including slang words, that were related to opiates. This also illustrates the importance of using the attention layer in the model to achieve explainable results. 
\begin{table}[htb]
	\centering
	\caption{Four examples showing the five most important words in each of them.}
	\label{tab:attention}
	\begin{tabular} {|p{6.5in}|}
		\hline
		\textbf{A post from an opioid user that contains the keyword ``opiates'':} \\ \hline
		\setlength{\fboxsep}{0pt} Before I've consumed any \colorbox{yellow}{drug} I've researched it $\dots$ I was tripping it was more \colorbox{yellow}{subtle} $\dots$ Besides that I know what \colorbox{yellow}{opiates} to do people $\dots$ but non lethal level and \colorbox{yellow}{oxy} was okay $\dots$ I've got \colorbox{yellow}{weed} to feel good and relax me.\\ \hline
		\textbf{A post from a non opioid user that contains the keyword ``opiates'':}
		 \\ \hline
		\setlength{\fboxsep}{0pt} While I \colorbox{yellow}{feel} \colorbox{yellow}{like} $\dots$ the amount \colorbox{yellow}{got} in my system $\dots$ his cup has \colorbox{yellow}{sub} in it and no \colorbox{yellow}{opiates} $\dots$ \\ \hline
		\textbf{A post from an opioid user that contains the slang word ``chocolate'':} \\ \hline
		\setlength{\fboxsep}{0pt} Again up to 600 mg \colorbox{yellow}{pretty} often although \dots \colorbox{yellow}{Feeling} like that $\dots$ if you take a \colorbox{yellow}{high} enough dose $\dots$ depends on the \colorbox{yellow}{type} \colorbox{yellow}{chocolate} more specifically $\dots$ \\ \hline
		\textbf{A post from a non opioid user that contains the slang word ``chocolate'':} \\ \hline
		\setlength{\fboxsep}{0pt} \colorbox{yellow}{Whenever} they mess up $\dots$ I have one \colorbox{yellow}{good} dBoy that $\dots$ The \colorbox{yellow}{opium} is a dreamy sedated \colorbox{yellow}{high} $\dots$ My \colorbox{yellow}{chocolate} milk came out of my nose when I laughed. \\ \hline
	\end{tabular}
\end{table}

\section{Discussion and Conclusion}

In this paper, we present a new application of an attention based bidirectional long short term memory model  to predict opioid users based on social media posts.  Our experimental results illustrate that it is feasible to learn the possible opioid users from their relevant Reddit posts with a reasonable accuracy and   the Att-BLSTM model outperforms traditional prediction approaches as well as the BLSTM model. The attention layer allows the model to capture important words that are relevant to the learning task, leading to a more explainable model. The model and the predictions can serve as the first step for further studies. For example, we can further obtain more posts from these opioid users by crawling Reddit and analyze the situation of these users, such as the sentiment of their posts,  any plan to quit drugs, and any difficulties they face. Once we can obtain their thoughts, problems, or difficulties, we may even involve social workers,  medical practitioners,  or governmental agencies to provide helps to them anonymously. In this direction, the AI backed algorithm can potentially lead to real social impact in battling the opioid epidemic.


There exists tremendous data on the web posted by anonymous users and the accumulation of such data on social media platforms increases at an accelerated pace.  Learning models and tools that can automatically extract useful knowledge are in great need. However, one practical barrier before one can apply existing or novel tools on such data is that the data is not in a learnable format: for example, it might need manual labeling before one can apply any classification tools. In this study, we construct our dataset by crawling from Reddit website and manually labeled each post/user via crowdsourcing, which takes quite some efforts. On the one hand, this dataset can serve as a gold standard  for future researchers who are interested in this topic to test their newly proposed methods. On the other hand, as a possible direction for future work, we will investigate the ideas of semi-supervised learning to obtain more labeled data automatically by taking advantage of the small seed data with labels. The field of language models has seen tremendous developments recently~\cite{Devlin2018,openai2023gpt4}. Investigation using newer and more powerful models on the problem is warranted.  

\section*{Acknowledgements}

The authors thank all the student researchers for their work in labeling the Reddit users.\vspace*{-12pt}

\section*{Funding}

This work has been supported in part by NSF CCF-2006780, IIS 2027667, and CCF 1815139. \vspace*{-12pt}
\\

\bibliographystyle{elsarticle-num}
\bibliography{bib}

\begin{thebibliography}{10}
\expandafter\ifx\csname url\endcsname\relax
  \def\url#1{\texttt{#1}}\fi
\expandafter\ifx\csname urlprefix\endcsname\relax\def\urlprefix{URL }\fi
\expandafter\ifx\csname href\endcsname\relax
  \def\href#1#2{#2} \def\path#1{#1}\fi

\bibitem{GladdenRMODonnellJMattsonCL2019}
R.~M. Gladden, J.~O’Donnell, C.~L. Mattson, P.~Seth, Changes in opioid-involved overdose deaths by opioid type and presence of benzodiazepines, cocaine, and methamphetamine—25 states, july--december 2017 to january--june 2018, Morbidity and Mortality Weekly Report 68~(34) (2019) 737.

\bibitem{Kalkman2019}
G.~A. Kalkman, C.~Kramers, R.~T. van Dongen, W.~van~den Brink, A.~Schellekens, Trends in use and misuse of opioids in the netherlands: a retrospective, multi-source database study, The Lancet Public Health 4~(10) (2019) e498--e505.

\bibitem{Yates2013}
A.~Yates, N.~Goharian, Adrtrace: detecting expected and unexpected adverse drug reactions from user reviews on social media sites, in: European Conference on Information Retrieval, Springer, 2013, pp. 816--819.

\bibitem{Liu2016}
J.~Liu, S.~Zhao, X.~Zhang, An ensemble method for extracting adverse drug events from social media, Artificial intelligence in medicine 70 (2016) 62--76.

\bibitem{Zhang2016}
Y.~Zhang, Y.~Fan, Y.~Ye, X.~Li, E.~L. Winstanley, Utilizing social media to combat opioid addiction epidemic: automatic detection of opioid users from twitter, in: AAAI Workshops, 2018.

\bibitem{Fan2017}
Y.~Fan, Y.~Zhang, Y.~Ye, X.~Li, W.~Zheng, Social media for opioid addiction epidemiology: Automatic detection of opioid addicts from twitter and case studies, in: Proceedings of the 2017 ACM on Conference on Information and Knowledge Management, 2017, pp. 1259--1267.

\bibitem{Fan2018}
Y.~Fan, Y.~Zhang, Y.~Ye, X.~Li, Automatic opioid user detection from twitter: Transductive ensemble built on different meta-graph based similarities over heterogeneous information network., in: IJCAI, 2018, pp. 3357--3363.

\bibitem{Mackey2017}
T.~K. Mackey, J.~Kalyanam, T.~Katsuki, G.~Lanckriet, Twitter-based detection of illegal online sale of prescription opioid, American journal of public health 107~(12) (2017) 1910--1915.

\bibitem{Yang2018}
Z.~Yang, L.~Nguyen, F.~Jin, Predicting opioid relapse using social media data, arXiv preprint arXiv:1811.12169 (2018).

\bibitem{Yao2020}
H.~Yao, S.~Rashidian, X.~Dong, H.~Duanmu, R.~N. Rosenthal, F.~Wang, {Detection of suicidality among opioid users on reddit: Machine learning-based approach}, Journal of Medical Internet Research (2020).
\newblock \href {https://doi.org/10.2196/15293} {\path{doi:10.2196/15293}}.

\bibitem{Zhou2016}
P.~Zhou, W.~Shi, J.~Tian, Z.~Qi, B.~Li, H.~Hao, B.~Xu, \href{https://aclanthology.org/P16-2034}{Attention-based bidirectional long short-term memory networks for relation classification}, in: K.~Erk, N.~A. Smith (Eds.), Proceedings of the 54th Annual Meeting of the Association for Computational Linguistics (Volume 2: Short Papers), Association for Computational Linguistics, Berlin, Germany, 2016, pp. 207--212.
\newblock \href {https://doi.org/10.18653/v1/P16-2034} {\path{doi:10.18653/v1/P16-2034}}.
\newline\urlprefix\url{https://aclanthology.org/P16-2034}

\bibitem{Ma2017}
F.~Ma, R.~Chitta, J.~Zhou, Q.~You, T.~Sun, J.~Gao, \href{https://doi.org/10.1145/3097983.3098088}{Dipole: Diagnosis prediction in healthcare via attention-based bidirectional recurrent neural networks}, in: Proceedings of the 23rd ACM SIGKDD International Conference on Knowledge Discovery and Data Mining, KDD '17, Association for Computing Machinery, New York, NY, USA, 2017, p. 1903–1911.
\newblock \href {https://doi.org/10.1145/3097983.3098088} {\path{doi:10.1145/3097983.3098088}}.
\newline\urlprefix\url{https://doi.org/10.1145/3097983.3098088}

\bibitem{LI2020}
C.~Li, Z.~Bao, L.~Li, Z.~Zhao, \href{https://www.sciencedirect.com/science/article/pii/S0306457319307204}{Exploring temporal representations by leveraging attention-based bidirectional lstm-rnns for multi-modal emotion recognition}, Information Processing \& Management 57~(3) (2020) 102185.
\newblock \href {https://doi.org/https://doi.org/10.1016/j.ipm.2019.102185} {\path{doi:https://doi.org/10.1016/j.ipm.2019.102185}}.
\newline\urlprefix\url{https://www.sciencedirect.com/science/article/pii/S0306457319307204}

\bibitem{Hochreiter1997}
S.~Hochreiter, J.~Schmidhuber, Long short-term memory, Neural computation 9~(8) (1997) 1735--1780.

\bibitem{Cheng2016}
J.~Cheng, L.~Dong, M.~Lapata, Long short-term memory-networks for machine reading, arXiv preprint arXiv:1601.06733 (2016).

\bibitem{Wang2016}
Y.~Wang, M.~Huang, X.~Zhu, L.~Zhao, Attention-based lstm for aspect-level sentiment classification, in: Proceedings of the 2016 conference on empirical methods in natural language processing, 2016, pp. 606--615.

\bibitem{Seo2016}
P.~H. Seo, Z.~Lin, S.~Cohen, X.~Shen, B.~Han, Progressive attention networks for visual attribute prediction, arXiv preprint arXiv:1606.02393 (2016).

\bibitem{Glove}
J.~Pennington, R.~Socher, C.~D. Manning, Glove: Global vectors for word representation, in: Proceedings of the 2014 conference on empirical methods in natural language processing (EMNLP), 2014, pp. 1532--1543.

\bibitem{Schuster1997}
M.~Schuster, K.~K. Paliwal, Bidirectional recurrent neural networks, IEEE transactions on Signal Processing 45~(11) (1997) 2673--2681.

\bibitem{Dea2017}
{United States Drug Enforcement Administration Houston Division}, Drug slang code words (2017).

\bibitem{Devlin2018}
J.~Devlin, M.~Chang, K.~Lee, K.~Toutanova, \href{http://arxiv.org/abs/1810.04805}{{BERT:} pre-training of deep bidirectional transformers for language understanding}, CoRR abs/1810.04805 (2018).
\newblock \href {http://arxiv.org/abs/1810.04805} {\path{arXiv:1810.04805}}.
\newline\urlprefix\url{http://arxiv.org/abs/1810.04805}

\bibitem{openai2023gpt4}
OpenAI, Gpt-4 technical report (2023).
\newblock \href {http://arxiv.org/abs/2303.08774} {\path{arXiv:2303.08774}}.

\end{thebibliography}

\end{document}